\pdfoutput=1
\documentclass[10pt,twocolumn]{article}
\usepackage[a4paper,top=20mm,bottom=22mm,left=16mm,right=16mm,columnsep=6mm]{geometry}
\usepackage[T1]{fontenc}
\usepackage{amsmath}
\usepackage{newtxtext,newtxmath}
\usepackage{bm}
\usepackage[numbers,sort&compress]{natbib}
\usepackage[raggedright]{titlesec}
\usepackage{caption}
\usepackage{graphicx}
\usepackage{booktabs}
\usepackage[dvipsnames]{xcolor}
\usepackage{placeins}
\usepackage{colortbl}
\usepackage{tabularx}
\usepackage[hyphens]{url}
\definecolor{paperrow}{gray}{0.95}
\newcommand{\papertablefont}{\renewcommand{\baselinestretch}{1}\fontsize{10}{12}\selectfont}
\newcommand{\papertablestyle}{%
  \papertablefont
  \setlength{\tabcolsep}{2pt}%
  \renewcommand{\arraystretch}{1.0}%
  \setlength{\heavyrulewidth}{0.6pt}%
  \setlength{\lightrulewidth}{0.35pt}%
}
\newcommand{\paperdensetablestyle}{\renewcommand{\baselinestretch}{1}\fontsize{8}{9.6}\selectfont\setlength{\tabcolsep}{1pt}\renewcommand{\arraystretch}{1.0}\setlength{\heavyrulewidth}{0.6pt}\setlength{\lightrulewidth}{0.35pt}}
\newcommand{\papertablehead}[1]{\textbf{#1}}
\newcommand{\papertwoline}[2]{\begin{tabular}[b]{@{}r@{}}\papertablehead{#1}\\\papertablehead{#2}\end{tabular}}

\usepackage[hidelinks,unicode]{hyperref}
\graphicspath{{images/}}

\renewenvironment{abstract}{\par\noindent\textbf{Abstract}\par\smallskip\noindent\ignorespaces}{\par}
\newenvironment{keyword}{\par\medskip\noindent\textbf{Keywords:}\ }{\par}
\newcommand{\sep}{\unskip; }
\begin{document}
\twocolumn[{%
\begin{center}
{\LARGE\bfseries SCR-Mamba: Skeleton-conditioned Mamba with cross-scale incremental refinement for efficient 3D human pose estimation\par}
\vspace{4mm}
{\large Jie Wang\textsuperscript{a,b}, Xuejing Gu\textsuperscript{*}, Congzhe Zhang, Zibo Zhao\par}
\vspace{2mm}
{\small \textsuperscript{a}College of Artificial Intelligence, North China University of Science and Technology, Tangshan 063210, China\par
\textsuperscript{b}Hebei Key Laboratory of Industrial Intelligent Perception, Tangshan 063210, China\par
\textsuperscript{*}Corresponding author: Xuejing Gu (\href{mailto:guxuejing@sina.com}{guxuejing@sina.com})\par}
\end{center}
\vspace{2mm}
{\small\begin{abstract}
Mamba-based methods for video-based 3D human pose estimation can model long sequences with linear complexity, but their selective state updates usually lack direct exploitation of skeletal relations. Meanwhile, stacking modeling blocks at the full temporal resolution leads to high computational overhead, while compressing the temporal sequence makes it difficult to retain per-frame details. To address these issues, we propose SCR-Mamba, whose core consists of Skeleton-Conditioned State Modeling (SCSM) and Cross-Scale Incremental Refinement (CSIR). SCSM generates a skeleton context from bone connections and left--right symmetry relations and generates the selective parameters of the spatial and temporal branches conditioned on it, so that structural relations are involved in the writing, propagation, and reading of state information. CSIR constructs multi-scale motion representations through content-adaptive aggregation and progressively refines the retained per-frame features with the increments produced by coarse-scale modeling, integrating long-range motion context while reducing full-resolution computation. On Human3.6M with Stacked Hourglass (SH) 2D detections and 243-frame inputs, SCR-Mamba-L achieves 36.8\,mm MPJPE (P1) with 7.72\,M parameters and 15.10\,G MACs, less than 1/7 of the computational cost of StructMamPose-L with similar accuracy; SCR-Mamba-X further reaches 36.6\,mm P1 and 30.9\,mm P-MPJPE (P2), achieving state-of-the-art performance. On MPI-INF-3DHP, SCR-Mamba-X achieves a percentage of correct keypoints (PCK) of 99.3\%, an area under the curve (AUC) of 88.1\%, and an MPJPE of 14.2\,mm.
\end{abstract}
}
\vspace{6mm}
}]
\section{Introduction}
\label{sec:introduction}
\begin{figure}[!b]
\centering
\includegraphics[width=\linewidth]{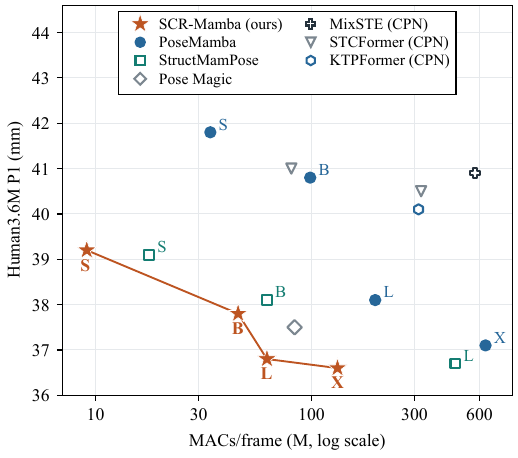}
\caption{Relationship between P1 and the per-frame computational cost (MACs/frame). The horizontal axis is on a logarithmic scale. All methods take 243 frames as input and predict the poses of the whole sequence. Methods marked with CPN in the legend use CPN inputs, and the others use SH inputs. Filled markers denote results under the unified counting rule, and hollow markers denote values reported in the original papers. S/B/L/X denote model scales.}
\label{fig:intro-tradeoff}
\end{figure}

Monocular 3D human pose estimation (HPE), which aims to recover the 3D coordinates of human joints from single-view images or videos, has become increasingly important in computer vision due to its wide range of applications in fields such as motion analysis, human--computer interaction, and behavior understanding. Benefiting from advances in 2D pose detection~\cite{Newell2016Hourglass,CPN2018}, current mainstream approaches adopt the 2D-to-3D lifting framework, which first detects 2D joints in images and then lifts them to 3D space through a deep network~\cite{SimpleBaseline2017,lu2025sama,PoseMoE2025}. However, owing to the lack of depth information in monocular observations, recovering 3D poses from 2D poses is inherently an ill-posed problem, because a single 2D projection can correspond to an infinite set of 3D poses. Therefore, effectively integrating human structure and cross-frame motion information while balancing estimation accuracy and computational efficiency remains a core challenge~\cite{KTPFormer2024,lu2025sama}.

\begin{figure*}[!t]
\centering
\includegraphics[width=\linewidth]{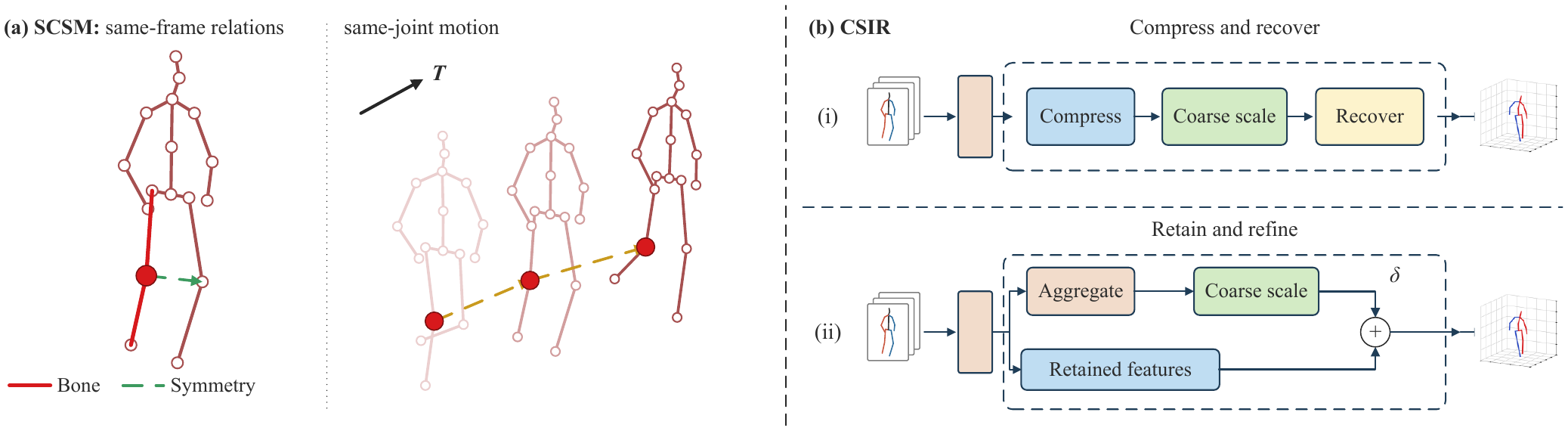}
\caption{Design motivation of SCR-Mamba. \textbf{(a)} SCSM: bone-connected and left--right symmetric joints in the same frame provide structural relations, and the cross-frame trajectory of the same joint describes motion changes. The target joint is marked in red, and $T$ denotes the temporal direction. \textbf{(b)} CSIR: (i) the network output is expanded to the original full temporal resolution after compression, as in HoT and H$_2$OT~\cite{li2024hot,li2026h2ot}; (ii) per-frame features are retained, and the increment $\delta$ produced by coarse-scale modeling is incorporated. Detailed structures are shown in Figs.~\ref{fig:architecture} and~\ref{fig:consistent-transfer}.}
\label{fig:intro-concept}
\end{figure*}

Transformer-based methods have achieved significant progress by leveraging the self-attention mechanism to model long-range spatio-temporal dependencies~\cite{zhang2022mixste,STCFormer2023}, but their computational complexity grows quadratically with the sequence length. As illustrated in Fig.~\ref{fig:intro-tradeoff}, the per-frame computational costs of MixSTE and KTPFormer both exceed 300\,M MACs. Mamba models long-range dependencies through selective state updates while maintaining linear computational complexity, providing a new avenue for efficient video-based pose estimation~\cite{gu2024mamba}. For example, PoseMamba learns spatio-temporal dependencies through bidirectional scanning~\cite{huang2025posemamba}. Recent studies further introduce structural information into state updates: SAMA fuses the skeleton topology at the level of joint features and hidden states and adaptively regulates the timescale according to joint motion characteristics~\cite{lu2025sama}; StructMamPose controls the writing and reading of state information through identity anchoring and enhances depth reasoning through internal view transformation~\cite{jiang2026structmampose}.

Although these methods introduce structural information into state updates, their effects concentrate on the fusion of features and hidden states or on the generation of a subset of the parameters. The step size, state writing, and state reading have not yet been jointly regulated by the same skeleton context. In practice, joint motion is closely related to the states of adjacent body parts. This requires not only the spatial branch to associate joints within the same frame but also the temporal branch to understand the motion of the same joint under different skeletal states, as shown in Fig.~\ref{fig:intro-concept}(a). On the other hand, local pose details and long-range motion context need to be jointly involved in per-frame prediction, whereas PoseMamba and SAMA mainly perform modeling at a fixed temporal resolution, which leads to high computational demand~\cite{huang2025posemamba,lu2025sama}. HoT and H$_2$OT reduce redundant computation by pruning pose tokens of redundant frames and recovering full-length tokens~\cite{li2024hot,li2026h2ot}, but their full-length outputs are mainly recovered from the retained representative tokens, and the features of the remaining frames are no longer involved in subsequent modeling. Skip connections in hierarchical networks retain the fine-scale features but add the complete upsampled coarse-scale features to them~\cite{Newell2016Hourglass}, so that the fine-scale information already contained in the coarse-scale features is superimposed again.

To address these challenges, we propose SCR-Mamba (\hspace{0pt}Skeleton-Conditioned Mamba with Cross-Scale Incremental Refinement), which consists of Skeleton-Conditioned State Modeling (SCSM) and Cross-Scale Incremental Refinement (CSIR). Specifically, SCSM aggregates bone-connection and left--right symmetry information from the current joint features to generate a skeleton context shared by the two branches, and uses it to generate the selective parameters of the spatial and temporal branches, so that structural relations directly affect state updates. While retaining per-frame features, CSIR constructs multi-scale motion representations through content-adaptive aggregation and takes the feature difference before and after coarse-scale modeling as the fine-scale update, as shown in Fig.~\ref{fig:intro-concept}(b). This design enables coarse-scale motion information to be incorporated into each frame representation in an incremental form, while placing the main computation on shorter sequences. As shown in Fig.~\ref{fig:intro-tradeoff}, SCR-Mamba attains estimation accuracy on par with recent advanced methods at a lower per-frame computational cost, achieving a favorable accuracy--efficiency trade-off.

The main contributions of this work are summarized as follows:
\begin{itemize}
\item We propose SCSM, which generates a skeleton context from bone connections and left--right symmetry relations and introduces it into the generation of selective parameters, so that structural relations are directly involved in state updates.
\item We propose CSIR, which transfers only the increments produced by coarse-scale modeling on the basis of retained per-frame features, thereby integrating multi-scale motion information while reducing full-resolution computation.
\item Based on these components, we develop SCR-Mamba at four scales (S, B, L, and X). Experiments on Human3.6M and MPI-INF-3DHP demonstrate that SCR-Mamba achieves state-of-the-art estimation accuracy while significantly reducing computational overhead.
\end{itemize}

The remainder of this paper is organized as follows. Section~\ref{sec:related-work} reviews related work. Section~\ref{sec:method} presents the proposed method. Section~\ref{sec:experiments} reports experimental results. Section~\ref{sec:conclusion} concludes the paper.

\section{Related work}
\label{sec:related-work}
\subsection{Structure-aware spatio-temporal modeling}
Video-based pose lifting requires jointly modeling joint relations and cross-frame motion, and existing studies mainly improve feature extraction in two respects: global dependencies and skeletal constraints. PoseFormer first utilizes the Transformer~\cite{Vaswani2017Transformer} to capture spatio-temporal dependencies~\cite{zheng2021poseformer}, and MixSTE and STCFormer further establish long-range connections through alternating or parallel spatio-temporal attention~\cite{zhang2022mixste,STCFormer2023}. MHFormer learns spatio-temporal representations of multiple plausible pose hypotheses to address the depth ambiguity of monocular estimation~\cite{Li2022MHFormer}. MotionAGFormer fuses attention and graph convolution in parallel to capture both global relations and local relations among adjacent joints~\cite{Mehraban2024MotionAGFormer}. Graph-based methods encode the skeleton topology through information propagation, such as SemGCN, GraFormer, and GLA-GCN~\cite{SemGCN2019,GraFormer2022,GLAGCN2023}, and GraphMLP further combines graph convolution with MLPs~\cite{Li2025GraphMLP}. DG-Net dynamically learns joint affinities according to the spatial distances and motion similarities between joints~\cite{Zhang2021DGNet}, while HDFormer exploits high-order bone and joint relations to handle overlapping joints and rapidly changing poses~\cite{Chen2023HDFormer}. KTPFormer introduces kinematics and trajectory priors into the attention parameters~\cite{KTPFormer2024}, whereas SDFormer applies anatomically grouped channel transformations~\cite{Zhong2026SDFormer}. These methods directly integrate human structure into dependency modeling, but the computational complexity of attention-based methods grows quadratically with the sequence length.

\subsection{Pose estimation with state space models}
Building upon the structured state space sequence model S4~\cite{Gu2022S4}, Mamba introduces input-dependent time-varying parameters and selective updates to model long sequences with linear complexity, providing an efficient basis for combining structural information with spatio-temporal dependencies~\cite{gu2024mamba}. Vision Mamba and VMamba further extend it to vision tasks, employing bidirectional SSM blocks and four-way 2D selective scanning, respectively~\cite{Zhu2024VisionMamba,Liu2024VMamba}. PoseMamba organizes spatio-temporal information through joint reordering and bidirectional scanning~\cite{huang2025posemamba}, Pose Magic integrates graph convolution with Mamba to capture both local skeletal correlations and global dependencies~\cite{PoseMagic2025}, and DBMambaPose alternately stacks decoupled spatial and temporal bidirectional blocks~\cite{Wang2026DBMambaPose}. HGMamba combines hypergraph convolution with Mamba to model local structures and global spatio-temporal features~\cite{Cui2025HGMamba}, whereas SasMamba constructs multi-scale structural representations through structure-aware spatio-temporal convolution and stride scanning~\cite{Cui2026SasMamba}. Gen3D-Mamba integrates forward, backward, and interval-sampled sequences and enhances generalization through hidden state pruning~\cite{Tao2026Gen3DMamba}; MHDPose combines temporal bidirectional Mamba with kinematic-aware spatial attention in a diffusion framework~\cite{Kappan2026MHDPose}. Recent studies have also introduced structural modeling into the state update process: SAMA fuses joint features and hidden states in the spatial branch and regulates the timescale with motion information in the temporal branch~\cite{lu2025sama}; StructMamPose injects spatio-temporal identities into the state parameter generation of horizontally flattened sequences, while its vertical scanning retains the standard update~\cite{jiang2026structmampose}. These designs introduce structural information at the level of features, hidden states, or identity injection. Unlike these designs, SCSM generates a single skeleton context from bone connections and left--right symmetry relations to jointly regulate the step size, state writing, and state reading of the spatial and temporal branches.

\subsection{Temporal compression and multi-scale modeling}
Long sequences contain rich motion information, but considerable redundancy also exists between adjacent frames. To reduce the sequence redundancy and computational cost, Strided Transformer and P-STMO shorten the sequence through progressive aggregation or temporal downsampling~\cite{StridedTransformer2023,PSTMO2022}, PoseFormerV2 and NanoHTNet utilize frequency-domain representations to retain long-range motion information~\cite{PoseFormerV22023,NanoHTNet2025}, and FreePose decomposes motion trajectories through the discrete wavelet transform~\cite{Liu2026FreePose}. For per-frame prediction, compression also needs to be combined with full-length recovery. HoT selects representative tokens and recovers the full-length sequence through cross-attention~\cite{li2024hot}, H$_2$OT further adopts multi-stage pruning~\cite{li2026h2ot}, and SCT compresses hidden features in the frequency domain and aggregates the upsampled features of each layer~\cite{Zheng2026SCT}. The full-length outputs of these methods are mainly recovered from the compressed representations. In contrast, inspired by coarse-grid correction in multigrid methods~\cite{Brandt1977}, CSIR retains fine-scale features and progressively updates each frame representation with the increments produced by coarse-scale modeling.

\section{Method}
\label{sec:method}
\subsection{Overall architecture}
\label{sec:overview}

\begin{figure*}[!t]
\centering
\includegraphics[width=\linewidth]{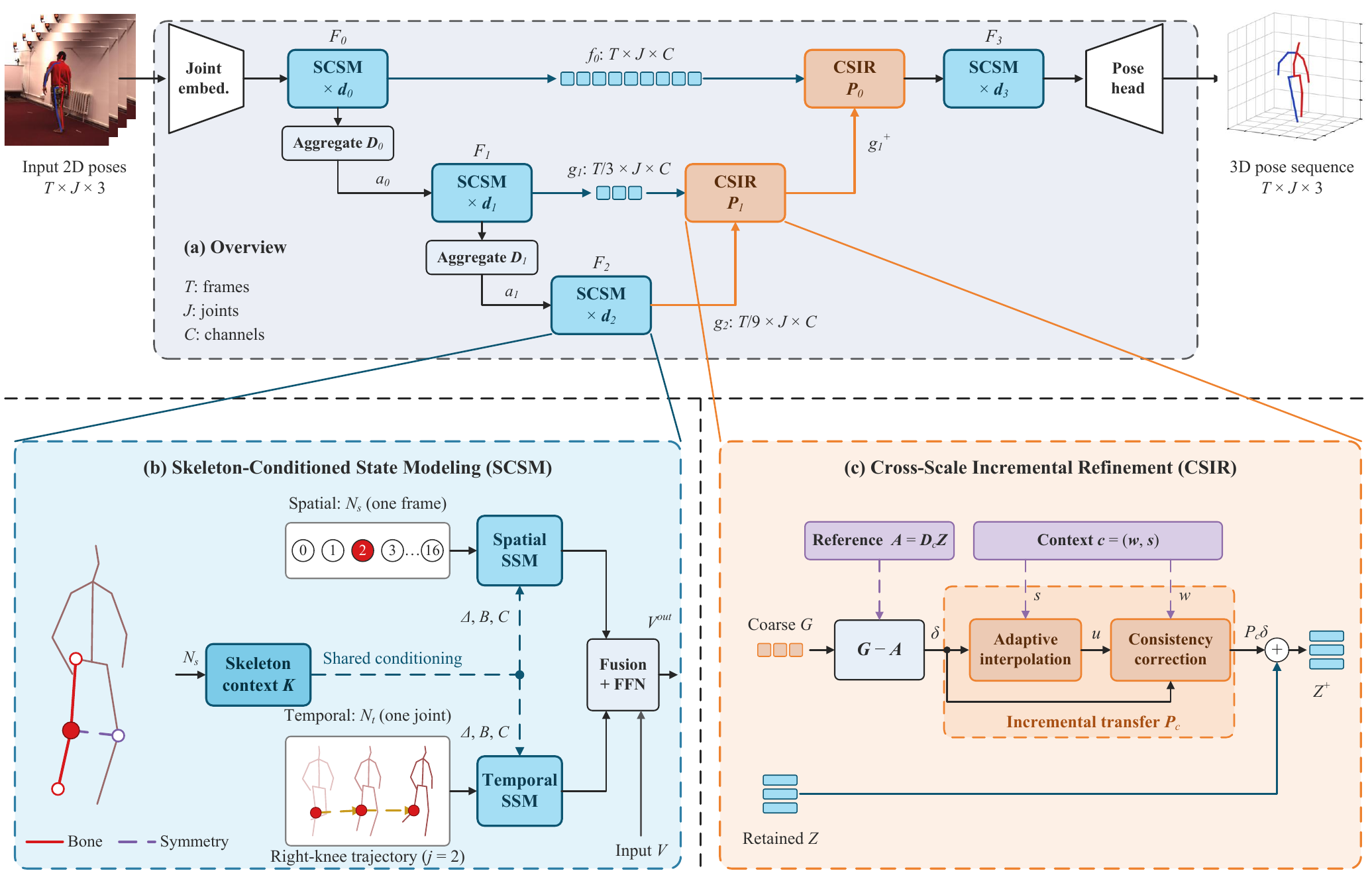}
\caption{Overall architecture of SCR-Mamba. \textbf{(a)} Stages $F_0$, $F_1$, and $F_2$ perform modeling at three temporal resolutions, CSIR progressively updates the retained features, and $F_3$ produces the output at the full resolution. \textbf{(b)} SCSM: bone connections and symmetry relations generate a shared skeleton context $K$, which guides the selective state updates of the spatial and temporal branches. \textbf{(c)} CSIR: the coarse-scale increment $\delta=G-A$ is extracted, processed by content-adaptive interpolation and consistent transfer, and added back to the fine-scale feature $Z$.}
\label{fig:architecture}
\end{figure*}

As illustrated in Fig.~\ref{fig:architecture}(a), SCR-Mamba takes 2D joint coordinates with their detection confidence scores $X\in\mathbb{R}^{T\times J\times 3}$ as input and predicts the 3D pose sequence $\hat Y\in\mathbb{R}^{T\times J\times 3}$, where $T$ and $J$ denote the numbers of frames and joints, respectively. The network is mainly composed of two modules, SCSM and CSIR: the former extracts joint and motion features at each temporal scale, and the latter updates the fine-scale representations with the increments of coarse-scale modeling.

First, a linear projection layer maps each joint in each frame to a $C$-dimensional feature space, spatial and temporal positional embeddings are added, and stage $F_0$ then extracts per-frame features. Subsequently, two levels of content-adaptive aggregation $D_0$ and $D_1$ progressively shorten the temporal length to $T/3$ and $T/9$, and stages $F_1$ and $F_2$ model motion context on the shorter sequences. Each stage consists of a stack of modeling blocks, each of which performs SCSM as shown in Fig.~\ref{fig:architecture}(b), with the numbers of joints $J$ and channels $C$ unchanged.

The network retains the fine-scale features before aggregation and the aggregation result at each level. CSIR in Fig.~\ref{fig:architecture}(c) generates updates from the difference between the coarse-scale modeling output and the aggregation result, first refining the intermediate scale and then updating the representations at the full temporal resolution. Finally, stage $F_3$ processes the updated per-frame features, and a regression head outputs the 3D poses. For a 243-frame input, the temporal lengths of the four stages are 243, 81, 27, and 243, respectively. The configurations of the four scales S/B/L/X are given in Section~\ref{sec:implementation}.

\subsection{Preliminaries}
\label{sec:preliminaries}
State space models (SSMs) map an input sequence $\bm x$ to an output sequence $\bm y$ through a hidden state $\bm h$. Their discrete form is
\begin{equation}
\bm h_k=\bar{\mathbf A}\bm h_{k-1}+\bar{\mathbf B}\bm x_k,\qquad
\bm y_k=\mathbf C\bm h_k,
\label{eq:ssm}
\end{equation}
where $\bar{\mathbf A}=\exp(\Delta\mathbf A)$, $\bar{\mathbf B}=(\Delta\mathbf A)^{-1}\bigl(\exp(\Delta\mathbf A)-\mathbf I\bigr)\Delta\mathbf B$, and $\Delta$ denotes the discretization step size. Mamba makes $\Delta$, $\mathbf B$, and $\mathbf C$ input-dependent, thereby regulating the writing, propagation, and reading of information according to the content~\cite{gu2024mamba}. These input-dependent quantities are referred to as selective parameters.

\subsection{Skeleton-conditioned state modeling (SCSM)}
\label{sec:spatiotemporal-block}

SCSM transforms the current skeleton features into a state update condition shared by the two branches, as shown in Fig.~\ref{fig:architecture}(b). Each branch takes its own features as the scan content, and the generation of its selective parameters is regulated by the skeleton context of the same frame.

\paragraph{Shared skeleton context}
Given the input $V\in\mathbb{R}^{\tau\times J\times C}$, where $\tau$ is the temporal length at the current scale, the two branches obtain $N_s=\operatorname{LN}_s(V)$ and $N_t=\operatorname{LN}_t(V)$ through layer normalization (LN), respectively. We aggregate neighborhood information on $N_s$ along bone connections and left--right symmetry relations, where the former links directly connected body parts and the latter links corresponding limbs. Omitting the temporal index and denoting the feature of joint $j$ as $\bm n_j$, the structural message and context are given by
\begin{equation}
\begin{aligned}
\bm m_j &=\sum_{\rho\in\{b,s\}}\alpha_\rho
 \sum_{k\in\mathcal N_\rho(j)}
 \bigl[\phi_r(\bm n_k-\bm n_j)+\phi_n(\bm n_k)\bigr],\\
\bm k_j &=\phi_o\bigl(\operatorname{GELU}(\bm m_j)\bigr),
\end{aligned}
\label{eq:structural-context}
\end{equation}
where $\mathcal N_b(j)$ and $\mathcal N_s(j)$ denote the bone and symmetry neighborhoods, respectively, $\phi_r,\phi_n,\phi_o$ are learnable affine mappings, and $\alpha_\rho$ is a learnable relation coefficient initialized to 1. The relative features describe the differences between the current joint and its adjacent body parts, and the neighbor features supplement the states of these parts. Performing this operation for each frame and each joint yields the shared context $K$. For example, the context of the right knee in Fig.~\ref{fig:architecture}(b) comes from the right hip, right ankle, and left knee.

\paragraph{Skeleton-conditioned dual-axis scanning}
We use $K$ to provide structural conditions for the generation of selective parameters in the two branches:
\begin{equation}
(\Delta_a,\mathbf B_a,\mathbf C_a)=\Phi_a(N_a+K),
\qquad a\in\{s,t\}.
\label{eq:structure-conditioned-control}
\end{equation}
Here, $\Delta_a$ is the state update step size, and $\mathbf B_a$ and $\mathbf C_a$ control the state input and readout, respectively; $\Phi_a$ includes the input projection, preprocessing, and parameter projections within the branch. The scan content and output gate are generated from $N_a$, and the skeleton context is only involved in determining how this content enters the state and is read out. In the spatial branch, this condition provides skeletal neighborhood information within the same frame; in the temporal branch, it supplements the joint feature at each time step with the structural state of the corresponding body part. The two branches can be formulated as
\begin{equation}
\begin{aligned}
S&=\mathcal S\bigl(N_s;\,N_s+K\bigr),\\
M&=\mathcal T\bigl(N_t;\,N_t+K\bigr),
\end{aligned}
\label{eq:parallel-axes}
\end{equation}
where the terms before and after the semicolon are the content input and the control context, respectively. The operator $\mathcal S$ scans the joint sequence of each frame, and $\mathcal T$ scans the cross-frame trajectory of each joint. Both follow bidirectional state space modeling~\cite{huang2025posemamba}, where the backward output is flipped back to the original order and added to the forward output.

\paragraph{Parallel fusion}
The spatial and temporal branches extract complementary features from the same input $V$, and these features are merged through learnable projections and then processed by a feed-forward network for channel interaction:
\begin{equation}
\begin{aligned}
\bar V &= V+SW_s^{\mathsf T}+MW_t^{\mathsf T},\\
V^{\mathrm{out}}&=\bar V+\operatorname{MLP}\bigl(\operatorname{LN}_f(\bar V)\bigr),
\end{aligned}
\label{eq:axis-fusion}
\end{equation}
where $W_s,W_t\in\mathbb{R}^{C\times C}$ are output projections, and MLP is a two-layer perceptron. This modeling block is repeated at each temporal scale with independently learned parameters.

\subsection{Cross-scale incremental refinement (CSIR)}
\label{sec:cross-scale-refinement}
\label{sec:temporal-aggregation}
\label{sec:consistent-transfer}

CSIR retains the fine-scale representations before each level of aggregation and progressively transfers only the increments produced by coarse-scale modeling to the fine scale, rather than recovering the fine-scale representations from coarse-scale features or superimposing coarse-scale features onto them. For a single scale transition, let $Z$ be the fine-scale feature, $A=D_cZ$ the aggregation result, and $G$ the coarse-scale modeling result. CSIR in Fig.~\ref{fig:architecture}(c) performs
\begin{equation}
\delta=G-D_cZ,\qquad Z^{+}=Z+P_c\delta,
\label{eq:single-level-update}
\end{equation}
where $D_c$ and $P_c$ denote the aggregation and increment transfer operators, respectively (the specific form of $P_c$ is given in Eq.~\eqref{eq:consistent-prolongation}), and $c$ is the content context saved during aggregation. The retention path carries per-frame features, while the increment path only supplements the changes produced by coarse-scale modeling. Aggregation and transfer share the content context, so that the construction of coarse-scale representations corresponds to the allocation of fine-scale updates.

\begin{figure}[!t]
\centering
\includegraphics[width=0.82\linewidth]{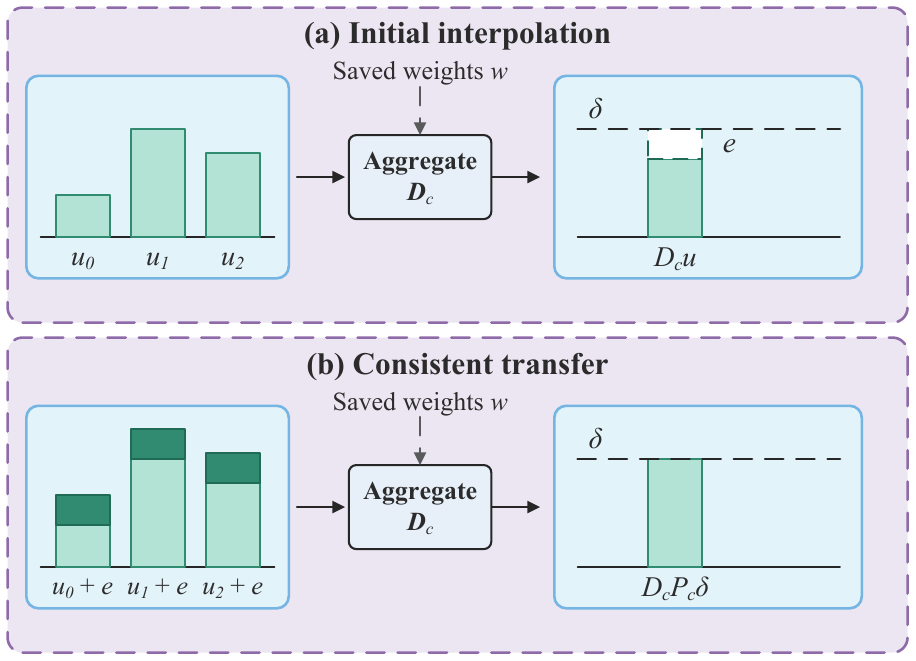}
\caption{Comparison between initial interpolation and consistent transfer. A group of three-frame updates is used for illustration, and both sides use the same aggregation weights $w$. \textbf{(a)} After re-aggregation, the initial interpolation $u$ leaves a residual $e=\delta-D_cu$ with respect to the coarse-scale target $\delta$. \textbf{(b)} The same residual (dark green) is added to the three frames, so that the re-aggregation result matches the target while the intra-group update differences are preserved. The dashed line indicates the target level. For visibility, the residual magnitude is enlarged and a positive residual is illustrated.}
\label{fig:consistent-transfer}
\end{figure}

\paragraph{Content-adaptive temporal aggregation}
Since different joints exhibit different motion changes within the same time segment, we generate aggregation weights according to the local features of each joint. Let $Z\in\mathbb{R}^{3Q\times J\times C}$ be divided into non-overlapping three-frame groups, and denote $\bm z_{q,r,j}=Z_{3q+r,j,:}$ with $r\in\{0,1,2\}$. The group mean is $\bm\mu_{q,j}=(1/3)\sum_r\bm z_{q,r,j}$, and the centered feature is $\bm v_{q,r,j}=\bm z_{q,r,j}-\bm\mu_{q,j}$. Root-mean-square normalization is performed along the intra-group temporal and channel dimensions:
\begin{equation}
\tilde{\bm v}_{q,r,j}=\frac{\bm v_{q,r,j}}
{\sqrt{\frac{1}{3C}\sum_{r'=0}^{2}\|\bm v_{q,r',j}\|_2^2+\varepsilon}},
\label{eq:group-normalization}
\end{equation}
where $\varepsilon=10^{-4}$. The aggregation weights are generated from the relative variation features as
\begin{equation}
\begin{aligned}
w_{q,r,j}&=\operatorname{softmax}_{r}\!\left(
\bm\theta_D^{\mathsf T}\tilde{\bm v}_{q,r,j}/\sqrt C\right),\\
(D_cZ)_{q,j,:}&=\sum_{r=0}^{2}w_{q,r,j}\bm z_{q,r,j}.
\end{aligned}
\label{eq:content-aggregation}
\end{equation}
Here, $\bm\theta_D\in\mathbb R^C$ is shared across temporal groups and joints, and the weights $w$ vary with the input content. Aggregation shortens the temporal length from $3Q$ to $Q$, while the numbers of joints and channels remain unchanged.

To enable the increment transfer to adapt to local content as well, we generate interpolation offsets from the same features:
\begin{equation}
s_{q,r,j}=(1/3)\tanh\!\left(
\bm\theta_U^{\mathsf T}\tilde{\bm v}_{q,r,j}/\sqrt C\right),
\label{eq:interpolation-context}
\end{equation}
where $\bm\theta_U\in\mathbb R^C$ is learned independently. Both vectors are initialized to zero, so that aggregation starts from the mean and interpolation starts from fixed three-phase interpolation. Each aggregation saves $c=(w,s)$, where $s$ is used to allocate fine-scale updates and $w$ is used to calibrate their aggregation results. The two levels of aggregation introduce $4C$ parameters in total.

\paragraph{Content-adaptive coarse-scale increment transfer}
The increment transfer first allocates the coarse-scale increment to fine temporal positions through interpolation $u=U_c\delta$. Within each three-frame group, the center frame receives the current coarse-scale increment, while the first and last frames combine it with the increments of the previous and next groups, respectively. The neighbor-group weights $\beta_{q,r,j}=1/3+s_{q,r,j}\in[0,2/3]$ vary with the input content. Specifically,
\begin{equation}
\begin{aligned}
\bm u_{q,0,j}&=(1-\beta_{q,0,j})\bm\delta_{q,j}
 +\beta_{q,0,j}\bm\delta_{q-1,j},\\
\bm u_{q,1,j}&=\bm\delta_{q,j},\\
\bm u_{q,2,j}&=(1-\beta_{q,2,j})\bm\delta_{q,j}
 +\beta_{q,2,j}\bm\delta_{q+1,j}.
\end{aligned}
\label{eq:adaptive-prolongation}
\end{equation}
Replicate padding is applied at both ends of the sequence, i.e., $\bm\delta_{-1,j}=\bm\delta_{0,j}$ and $\bm\delta_{Q,j}=\bm\delta_{Q-1,j}$.

Content-adaptive interpolation determines the update differences within each group, but these updates may deviate from the target increment $\delta$ after being aggregated with the original weights, as shown in Fig.~\ref{fig:consistent-transfer}(a). To make the two scales express the same modeling update, we further construct a closed-form correction to obtain the consistent transfer operator $P_c$.

\paragraph{Consistent transfer}
As shown in Fig.~\ref{fig:consistent-transfer}(b), the aggregation residual $e=\delta-D_cu$ is computed with the saved $w$ and distributed to the corresponding three frames by the repetition operator $R$:
\begin{equation}
P_c\delta=U_c\delta+R\bigl(\delta-D_cU_c\delta\bigr).
\label{eq:consistent-prolongation}
\end{equation}
Since the aggregation weights of each group sum to 1, $D_cR=I$ holds for a given content context $c$, and thus
\begin{equation}
\begin{aligned}
D_cP_c\delta&=D_cU_c\delta+\delta-D_cU_c\delta=\delta,\\
D_cZ^{+}&=D_cZ+\delta=G.
\end{aligned}
\label{eq:updated-feature-consistency}
\end{equation}
Therefore, the updated fine-scale representation is consistent with the coarse-scale result under the original aggregation rule. Meanwhile, the three frames of the same joint within a group receive the same residual offset, so that the intra-group differences formed by the initial interpolation are preserved. This correction reuses the content context and requires no additional parameters or iterative solving. Moreover, Eq.~\eqref{eq:single-level-update} can be rewritten as $Z^{+}=(I-P_cD_c)Z+P_cG$. Since $D_c(I-P_cD_c)=0$, the former term contains the fine-scale details that are invisible under the original aggregation rule, and the latter term transfers the coarse-scale result.

In the network, the above refinement is performed recursively from coarse to fine. For the $\ell$-th level of transfer, the saved content context is denoted by $c_\ell$, and the corresponding aggregation and transfer operators are written as $D_\ell$ and $P_\ell$, respectively. Let $f_0$ be the output of stage $F_0$, $a_0=D_0f_0$ and $g_1=F_1(a_0)$ be the input and output at the intermediate scale, and $a_1=D_1g_1$ and $g_2=F_2(a_1)$ be those at the coarsest scale. Then
\begin{equation}
\begin{aligned}
g_1^{+}&=g_1+P_1(g_2-a_1;c_1),\\
f_0^{+}&=f_0+P_0(g_1^{+}-a_0;c_0).
\end{aligned}
\label{eq:recursive-increment}
\end{equation}
The update at the coarsest scale is first incorporated into $g_1$ and then transferred to $f_0$ together with the modeling increment at the intermediate scale. The resulting per-frame representation is fed into $F_3$ and the regression head to output the 3D poses.

\subsection{Training objective}
\label{sec:training-objective}
SCR-Mamba is supervised at the full temporal resolution. Following MixSTE and PoseMamba~\cite{zhang2022mixste,huang2025posemamba}, we adopt the position, scale-aligned, velocity, and weighted smoothness losses. The annotations take the pelvis as the origin, and the losses are computed in the normalized coordinate space. The overall objective is defined as
\begin{equation}
\begin{aligned}
\mathcal L={}&\mathcal L_{\mathrm{pos}}+0.5\mathcal L_{\mathrm{scale}}\\
&+20\mathcal L_{\mathrm{vel}}+0.5\mathcal L_{\mathrm{smooth}},
\end{aligned}
\label{eq:training-objective}
\end{equation}
where $\mathcal L_{\mathrm{pos}}$ is the average Euclidean distance between the predicted and ground-truth joints, and $\mathcal L_{\mathrm{scale}}$ computes this distance after fitting a global scale for each frame by least squares. Moreover, $\mathcal L_{\mathrm{vel}}$ measures the Euclidean distance between the predicted and ground-truth displacements of adjacent frames, and $\mathcal L_{\mathrm{smooth}}$ weights the squared first-order differences of the predictions by joint and averages them over time, joints, and coordinate components. In the smoothness term, the weights are 2.5 for the knees and ankles, 1.5 for the head and shoulders, 4 for the elbows and wrists, and 1 for the remaining joints. The latter two terms are computed over $T-1$ pairs of adjacent frames. The network is optimized in an end-to-end fashion.

\section{Experiments}
\label{sec:experiments}
This section first introduces the evaluation protocols and implementation details, and then compares SCR-Mamba with existing methods in terms of accuracy, computational cost, and inference speed. Subsequently, ablation studies analyze the effects of SCSM and CSIR, followed by qualitative analysis on specific poses.
\subsection{Datasets and evaluation protocols}
\label{sec:evaluation-protocol}
\paragraph{Human3.6M}
For Human3.6M~\cite{ionescu2014human36m}, we follow the standard split, where subjects S1, S5, S6, S7, and S8 are used for training, while subjects S9 and S11 are reserved for testing. The inputs are 2D joints and their confidence scores detected by Stacked Hourglass (SH)~\cite{Newell2016Hourglass}, and each clip contains 243 frames with 17 joints per frame. The models with ground-truth 2D inputs ($\mathrm{P1}^{\mathrm{GT}}$) replace the detected results with annotated 2D joints, and are trained and evaluated separately from those with SH inputs. We report the mean per-joint position error (MPJPE, denoted P1) after root (pelvis) alignment and the error after similarity transformation alignment (P2), both in millimeters~\cite{pavllo2019videopose,huang2025posemamba}. The ablation studies also report the mean per-joint velocity error (MPJVE), i.e., the average Euclidean norm of the difference between the predicted and ground-truth displacements of adjacent frames, in mm/frame~\cite{pavllo2019videopose}.

\paragraph{MPI-INF-3DHP}
For MPI-INF-3DHP~\cite{mehta2017mpi}, we follow the official training and test splits, take 81-frame ground-truth 2D joints as input, and evaluate on TS1--TS6. Evaluation is performed in pelvis-relative coordinates: MPJPE uses 17 joints, the percentage of correct keypoints (PCK) uses the first 14 official joints with a 150\,mm threshold. The area under the curve (AUC) is computed as the mean PCK over 31 thresholds from 0 to 150\,mm at an interval of 5\,mm.

\subsection{Implementation details}
\label{sec:implementation}
By adjusting the channel width and stage depths, we construct four configurations, SCR-Mamba-S, B, L, and X, as summarized in Table~\ref{tab:model-configurations}. The inner channels of both the spatial and temporal branches are $2C$, with state dimensions of 4 and 16, respectively; the depthwise convolution kernel size of the temporal branch is 3, and the hidden dimension of the feed-forward network is $2C$.

SCR-Mamba-L and the ablation configurations are trained from random initialization for 40 epochs using the AdamW optimizer~\cite{AdamW2019} with a batch size of 4, an initial learning rate of $5\times10^{-4}$, and a weight decay of $0.01$. The learning rate is decayed by a factor of 0.99 after each epoch. Horizontal flipping is applied as augmentation for both training and testing, and the decay rate of the exponential moving average (EMA) is set to 0.9998. The S and B configurations use the same number of epochs, initial learning rate, learning rate decay, weight decay, and EMA settings.

\begin{table}[!t]
\centering
\caption{Model configurations. $C$: channel width. Depth: stage depths. Params and MACs are counted over the whole network.}
\label{tab:model-configurations}
\papertablestyle
\begin{tabularx}{\columnwidth}{@{}Xrcrr@{}}\toprule
\papertablehead{Model} & $C$ & \papertablehead{Depth} & \papertablehead{Params (M)} & \papertablehead{MACs (G)} \\ \midrule
SCR-Mamba-S & 64 & $[1,4,4,1]$ & 1.09 & 2.21 \\
SCR-Mamba-B & 128 & $[2,4,2,2]$ & 3.87 & 11.11 \\
SCR-Mamba-L & 128 & $[2,8,8,2]$ & 7.72 & 15.10 \\
SCR-Mamba-X & 192 & $[2,8,8,2]$ & 16.62 & 32.05 \\ \bottomrule
\end{tabularx}
\end{table}

SCR-Mamba-X increases the channel width to 192 and is trained with an initial learning rate of $4\times10^{-4}$, with the other settings the same as those of L. On MPI-INF-3DHP, SCR-Mamba-X is trained from random initialization with a batch size of 4 and the same optimizer, learning rate, and loss coefficients as on Human3.6M.

\subsection{Comparison with state-of-the-art methods}
\label{sec:main-comparison}
\begin{table*}[!t]
\centering
\caption{Comparison of accuracy and model cost on Human3.6M. All methods take 243 frames as input and output the whole sequence. P1/P2 use the listed 2D detectors, and $\mathrm{P1}^{\mathrm{GT}}$ uses ground-truth 2D joints as input. Errors are in mm. Our models are denoted by bold names. The lowest and second-lowest displayed errors are in bold and underlined, respectively, and ties are marked equally.}
\label{tab:main-comparison}

\paperdensetablestyle
\begin{tabularx}{\linewidth}{@{}l>{\hsize=2.7\hsize\raggedright\arraybackslash}Xc>{\hsize=0.85\hsize\raggedleft\arraybackslash}X>{\hsize=0.75\hsize\raggedleft\arraybackslash}X>{\hsize=1.05\hsize\raggedleft\arraybackslash}X>{\hsize=0.55\hsize\raggedleft\arraybackslash}X>{\hsize=0.55\hsize\raggedleft\arraybackslash}X>{\hsize=0.55\hsize\raggedleft\arraybackslash}X@{}}
\toprule
\papertablehead{Size} & \papertablehead{Method} & \papertablehead{2D} & \papertwoline{Params}{(M)} & \papertwoline{MACs}{(G)} & \papertwoline{MACs/frame}{(M)} & \papertablehead{P1}$\downarrow$ & \papertablehead{P2}$\downarrow$ & $\mathbf{P1}^{\mathbf{GT}}\!\downarrow$ \\
\midrule
S & PoseMamba-S~\cite{huang2025posemamba} {\scriptsize(AAAI'25)} & SH & 0.86 & 3.58 & 14.76 & 41.8 & 35.0 & 20.0 \\
 & SAMA-S~\cite{lu2025sama} {\scriptsize(ICCV'25)} & SH & 1.1 & 3.9 & 16.04 & 40.6 & 34.0 & 20.2 \\
 & StructMamPose-S~\cite{jiang2026structmampose} {\scriptsize(ICML'26)} & SH & 1.2 & 4.3 & 17.69 & 39.1 & 32.9 & 18.8 \\
 & \textbf{SCR-Mamba-S} & SH & 1.09 & 2.21 & 9.10 & 39.2 & 32.8 & 17.3 \\
\addlinespace[4pt]
B & STCFormer~\cite{STCFormer2023} {\scriptsize(CVPR'23)} & CPN & 4.75 & 19.6$^{\ddagger}$ & 80.65 & 41.0 & 32.0 & 21.3$^{\dagger}$ \\
 & PoseMamba-B {\scriptsize(AAAI'25)} & SH & 3.35 & 13.94 & 57.37 & 40.8 & 34.3 & 16.8 \\
 & SAMA-B {\scriptsize(ICCV'25)} & SH & 3.3 & 11.7 & 48.14 & 37.7 & 32.0 & 13.6 \\
 & SDFormer-B~\cite{Zhong2026SDFormer} {\scriptsize(PR'26)} & SH & 7.5 & 22.3 & 91.76 & 37.8 & 31.7 & 15.7 \\
 & StructMamPose-B {\scriptsize(ICML'26)} & SH & 4.9 & 15.1 & 62.13 & 38.1 & 32.3 & 14.2 \\
 & \textbf{SCR-Mamba-B} & SH & 3.87 & 11.11 & 45.72 & 37.8 & 31.8 & 15.6 \\
\addlinespace[4pt]
L & MixSTE~\cite{zhang2022mixste} {\scriptsize(CVPR'22)} & CPN & 33.6$^{\dagger}$ & 139.0$^{\dagger}$ & 572.01 & 40.9 & 32.6 & 21.6$^{\dagger}$ \\
 & STCFormer-L {\scriptsize(CVPR'23)} & CPN & 18.91 & 78.2$^{\ddagger}$ & 321.81 & 40.5 & 31.8 & --- \\
 & KTPFormer~\cite{KTPFormer2024} {\scriptsize(CVPR'24)} & CPN & 35.2$^{\ddagger}$ & 76.1$^{\ddagger}$ & 313.16 & 40.1 & 31.9 & 19.0$^{\ddagger}$ \\
 & MotionBERT (scratch)~\cite{zhu2023motionbert} {\scriptsize(ICCV'23)} & SH & 42.3$^{\ddagger}$ & 174.8$^{\ddagger}$ & 719.34 & 39.2 & 32.9$^{\ddagger}$ & 17.8$^{\dagger}$ \\
 & TCPFormer~\cite{TCPFormer2025} {\scriptsize(AAAI'25)} & SH & 35.1 & 109.2 & 449.38 & 37.9 & 31.7 & 15.5 \\
 & Pose Magic~\cite{PoseMagic2025} {\scriptsize(AAAI'25)} & SH & 14.42 & 20.29$^{\ddagger}$ & 83.49 & 37.5 & --- & --- \\
 & PoseMamba-L {\scriptsize(AAAI'25)} & SH & 6.71 & 27.88 & 114.73 & 38.1 & 32.5 & 15.6 \\
 & SAMA-L {\scriptsize(ICCV'25)} & SH & 17.3 & 53.2 & 218.93 & 36.9 & 31.3 & \textbf{11.9} \\
 & SCT~\cite{Zheng2026SCT} {\scriptsize(PR'26)} & SH & 42.6 & 58.9 & 242.38 & 37.7 & 31.7 & 15.4 \\
 & FreePose-L~\cite{Liu2026FreePose} {\scriptsize(PR'26)} & SH & 11.8 & 33.5 & 137.86 & 37.8 & 32.2 & 14.2 \\
 & StructMamPose-L {\scriptsize(ICML'26)} & SH & 31.1 & 112.7 & 463.78 & \underline{36.7} & \underline{31.1} & 13.4 \\
 & \textbf{SCR-Mamba-L} & SH & 7.72 & 15.10 & 62.17 & 36.8 & \underline{31.1} & 12.7 \\
\addlinespace[4pt]
X & PoseMamba-X {\scriptsize(AAAI'25)} & SH & 26.53 & 109.90 & 452.30 & 37.1 & 31.5 & 14.8 \\
 & \textbf{SCR-Mamba-X} & SH & 16.62 & 32.05 & 131.93 & \textbf{36.6} & \textbf{30.9} & \underline{12.5} \\
\bottomrule
\end{tabularx}
\par\smallskip
\parbox{\linewidth}{SH: Stacked Hourglass~\cite{Newell2016Hourglass}; CPN: Cascaded Pyramid Network~\cite{CPN2018}. Scales follow the naming of each model series, and MACs/frame denotes the average computational cost per output frame. $\dagger$ and $\ddagger$ denote values quoted from Table~1 of PoseMamba and SAMA~\cite{huang2025posemamba,lu2025sama}, respectively; the other comparison values are taken from the original papers. MotionBERT adopts the training-from-scratch configuration. For our models, the parameters of the whole network and the MACs including selective scanning are counted; without scanning, the MACs of SCR-Mamba-S/B/L/X are about 1.85, 10.03, 13.64, and 29.85\,G, respectively. The costs of the compared methods follow the counts reported in the literature. Dashes denote unreported results.}
\end{table*}

\paragraph{Human3.6M}
Table~\ref{tab:main-comparison} compares representative methods under the setting of 243-frame input and whole-sequence output. To ensure a fair comparison, we report only the results of models trained without additional pre-training on external data, and mark the SH and CPN inputs separately. The following analysis mainly focuses on methods with SH inputs.

With SH inputs, SCR-Mamba-L achieves P1/P2 errors of 36.8/31.1\,mm. Its P1 is 1.0, 1.0, 0.9, and 1.1\,mm lower than those of FreePose-L, SDFormer-B, SCT, and TCPFormer, respectively, and its P2 is 1.1, 0.6, 0.6, and 0.6\,mm lower. Its accuracy is comparable to that of SAMA-L (36.9/31.3\,mm) and StructMamPose-L (36.7/31.1\,mm). In terms of model cost, compared with the reported value of SDFormer-B, SCR-Mamba-L reduces the MACs from 22.3\,G to 15.10\,G. Compared with SAMA-L and StructMamPose-L of similar accuracy, SCR-Mamba-L uses less than 1/2 and 1/4 of their parameters and less than 1/3 and 1/7 of their MACs, respectively. SCR-Mamba-X further achieves P1/P2 errors of 36.6/30.9\,mm, reaching state-of-the-art performance.

The smaller configurations also achieve a favorable balance between accuracy and efficiency. With 2.21\,G and 11.11\,G MACs, SCR-Mamba-S and SCR-Mamba-B achieve P1 errors of 39.2\,mm and 37.8\,mm, respectively, and their computational costs are lower than the reported values of StructMamPose and SAMA at the same scales. With ground-truth 2D inputs, SCR-Mamba-X achieves a P1 of 12.5\,mm, the second-best result in the table.

\begin{table*}[!t]
\centering
\caption{P1 and P2 of each action on Human3.6M (mm, same setting as Table~\ref{tab:main-comparison}). Avg. denotes the average error over the 15 actions. The lowest and second-lowest displayed errors in each column are in bold and underlined, respectively, and our model is denoted by a bold name.}
\label{tab:action-comparison}
\paperdensetablestyle
\begin{tabularx}{\linewidth}{@{}l*{16}{>{\centering\arraybackslash}X}@{}}
\toprule
\papertablehead{Method} & \papertablehead{Dir.} & \papertablehead{Disc.} & \papertablehead{Eat} & \papertablehead{Greet} & \papertablehead{Phone} & \papertablehead{Photo} & \papertablehead{Pose} & \papertablehead{Pur.} & \papertablehead{Sit} & \papertablehead{SitD.} & \papertablehead{Smoke} & \papertablehead{Wait} & \papertablehead{WalkD} & \papertablehead{Walk} & \papertablehead{WalkT} & \papertablehead{Avg.} \\
\midrule
\multicolumn{17}{@{}l}{\textit{P1}} \\
MixSTE~\cite{zhang2022mixste} & 37.6 & 40.9 & 37.3 & 39.7 & 42.3 & 49.9 & 40.1 & 39.8 & 51.7 & 55.0 & 42.1 & 39.8 & 41.0 & 27.9 & 27.9 & 40.9 \\
STCFormer~\cite{STCFormer2023} & 39.6 & 41.6 & 37.4 & 38.8 & 43.1 & 51.1 & 39.1 & 39.7 & 51.4 & 57.4 & 41.8 & 38.5 & 40.7 & 27.1 & 28.6 & 41.0 \\
KTPFormer~\cite{KTPFormer2024} & 37.3 & 39.2 & 35.9 & 37.6 & 42.5 & 48.2 & 38.6 & 39.0 & 51.4 & 55.9 & 41.6 & 39.0 & 40.0 & 27.0 & 27.4 & 40.1 \\
MotionBERT~\cite{zhu2023motionbert} & 36.3 & 38.7 & 38.6 & 33.6 & 42.1 & 50.1 & 36.2 & 35.7 & 50.1 & 56.6 & 41.3 & 37.4 & 37.7 & \underline{25.6} & 26.5 & 39.2 \\
PoseMamba-L~\cite{huang2025posemamba}$^{\dagger}$ & 36.5 & \underline{37.9} & 37.5 & 33.5 & \textbf{39.6} & 47.7 & 36.5 & \underline{34.0} & \textbf{47.8} & 54.5 & 40.2 & 36.3 & 36.7 & 26.4 & 26.9 & 38.1 \\
SCT~\cite{Zheng2026SCT} & \underline{35.8} & 38.3 & \underline{35.7} & 32.2 & \underline{39.7} & 47.3 & \underline{36.0} & 34.5 & 50.0 & \underline{52.9} & \textbf{39.4} & 36.8 & \underline{35.7} & \textbf{25.2} & \textbf{25.3} & \underline{37.7} \\
FreePose-L~\cite{Liu2026FreePose} & \textbf{35.0} & \underline{37.9} & 36.5 & \underline{31.7} & 40.6 & \underline{46.7} & 36.8 & 34.7 & 48.5 & 53.4 & 39.8 & \underline{35.8} & 36.0 & 26.2 & 26.7 & 37.8 \\
StructMamPose-S~\cite{jiang2026structmampose} & 37.8 & 39.9 & 38.0 & 33.5 & 40.8 & 47.4 & 37.9 & 37.0 & 50.6 & 54.2 & 40.7 & 37.3 & 37.2 & 27.6 & 27.0 & 39.1 \\
\textbf{SCR-Mamba-L} & \underline{35.8} & \textbf{37.5} & \textbf{35.5} & \textbf{31.5} & 39.8 & \textbf{45.5} & \textbf{34.7} & \textbf{33.0} & \underline{47.9} & \textbf{50.4} & \underline{39.5} & \textbf{35.1} & \textbf{34.6} & \textbf{25.2} & \underline{25.7} & \textbf{36.8} \\
\midrule
\multicolumn{17}{@{}l}{\textit{P2}} \\
MixSTE~\cite{zhang2022mixste} & 30.8 & 33.1 & 30.3 & 31.8 & 33.1 & 39.1 & 31.1 & 30.5 & 42.5 & \underline{44.5} & \underline{34.0} & 30.8 & 32.7 & 22.1 & 22.9 & 32.6 \\
STCFormer~\cite{STCFormer2023} & \textbf{29.5} & 33.2 & 30.6 & 31.0 & \underline{33.0} & 38.0 & 30.4 & 29.4 & 41.8 & 45.2 & \textbf{33.6} & \underline{29.5} & 31.6 & \textbf{21.3} & 22.6 & 32.0 \\
KTPFormer~\cite{KTPFormer2024} & 30.1 & 32.3 & \textbf{29.6} & 30.8 & \textbf{32.3} & \textbf{37.3} & 30.0 & 30.2 & \underline{41.0} & 45.3 & \textbf{33.6} & 29.9 & 31.4 & 21.5 & 22.6 & 31.9 \\
MotionBERT~\cite{zhu2023motionbert}$^{\dagger}$ & 30.8 & 32.8 & 32.4 & 28.7 & 34.3 & 38.9 & 30.1 & 30.1 & 42.5 & 49.7 & 36.0 & 30.8 & 31.8 & 22.1 & 23.0 & 32.9 \\
PoseMamba-L~\cite{huang2025posemamba}$^{\dagger}$ & 31.0 & 32.2 & 32.5 & 28.5 & 33.1 & 38.6 & 30.5 & \underline{29.2} & 41.3 & 47.8 & 34.9 & 30.2 & 31.7 & 22.7 & 23.5 & 32.5 \\
SCT~\cite{Zheng2026SCT} & \underline{29.7} & \underline{31.6} & 30.6 & 27.2 & 33.1 & 37.6 & \underline{29.6} & 29.9 & 41.6 & 46.6 & 34.5 & 29.9 & \underline{30.6} & \underline{21.4} & \textbf{21.8} & \underline{31.7} \\
FreePose-L~\cite{Liu2026FreePose} & 30.3 & 32.2 & 31.6 & \underline{27.1} & 34.0 & 37.5 & 30.6 & 29.8 & 41.2 & 47.0 & 35.0 & 29.8 & 31.4 & 22.2 & 22.9 & 32.2 \\
\textbf{SCR-Mamba-L} & 30.0 & \textbf{31.3} & \underline{30.1} & \textbf{26.7} & 33.1 & \underline{37.4} & \textbf{28.5} & \textbf{28.9} & \textbf{39.8} & \textbf{44.3} & 34.2 & \textbf{28.7} & \textbf{29.9} & \textbf{21.3} & \underline{22.1} & \textbf{31.1} \\
\bottomrule
\end{tabularx}
\par\smallskip
{\footnotesize\parbox{\linewidth}{MixSTE, STCFormer, and KTPFormer use CPN, and the others use SH; MotionBERT adopts the training-from-scratch configuration. The comparison values and their Avg. are taken from the original papers. $\dagger$ denotes that the per-action results are not reported in the original paper and are computed from the official public weights on the same test inputs. StructMamPose only releases the per-action P1 of its S version.}}
\end{table*}

Table~\ref{tab:action-comparison} further reports the P1 and P2 of each action. With the same SH inputs, the P1 of SCR-Mamba-L is lower than that of MotionBERT on all 15 actions, with reductions of 6.2 and 4.6\,mm on SittingDown and Photo, respectively, and lower than that of FreePose-L on 14 actions. Its P1 and P2 are the lowest among all compared methods, including those with CPN inputs, on 10 and 9 of the 15 actions, respectively.

\begin{figure*}[t]
\centering
\includegraphics[width=\linewidth]{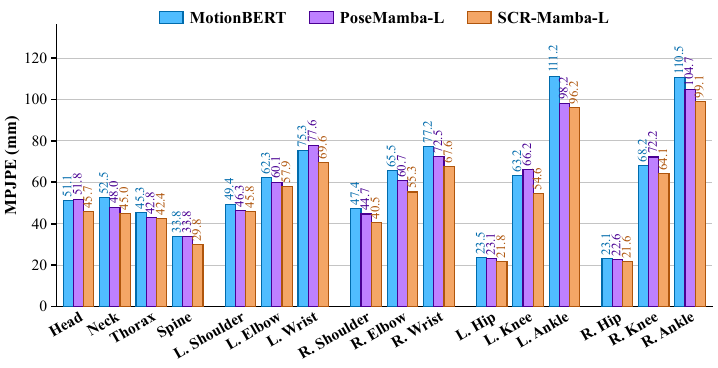}
\caption{Per-joint MPJPE comparison on the SittingDown action of Human3.6M (mm, SH input). L. and R. denote the left and right sides, respectively. The results of MotionBERT and PoseMamba-L are computed from their official public weights on the same test inputs.}
\label{fig:joint-improvements}
\end{figure*}

\begin{figure*}[!t]
\centering
\includegraphics[width=\linewidth]{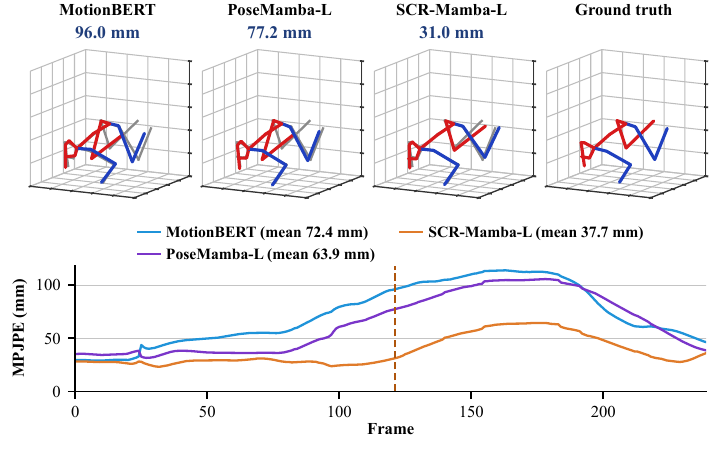}
\caption{Frame-wise MPJPE of a 240-frame clip from S11 SittingDown (camera 2) with large differences among the methods. Top: 3D poses and MPJPE of the frame marked by the dashed line, where blue and red denote the left and right limbs, respectively, and gray denotes the ground truth. Bottom: frame-wise error curves of the three methods, with the clip averages given in the legend.}
\label{fig:framewise}
\end{figure*}

Fig.~\ref{fig:joint-improvements} compares the per-joint errors on SittingDown, where the largest difference appears. SCR-Mamba-L achieves lower errors than MotionBERT and PoseMamba-L on all non-root joints, where the error of the left ankle is reduced by 15.0\,mm compared with MotionBERT and that of the left knee is reduced by 11.6\,mm compared with PoseMamba-L. Fig.~\ref{fig:framewise} further presents a 240-frame clip from S11 SittingDown with large differences among the methods, on which the average MPJPEs of the three methods are 72.4, 63.9, and 37.7\,mm, respectively.

\paragraph{MPI-INF-3DHP}
To evaluate performance on diverse scenarios, we conduct experiments on MPI-INF-3DHP, which covers both indoor and outdoor scenes. As shown in Table~\ref{tab:mpi-comparison}, SCR-Mamba-X achieves a PCK of 99.3\%, an AUC of 88.1\%, and an MPJPE of 14.2\,mm. Its PCK is the highest in the table, its MPJPE is the second-best, 0.1\,mm higher than that of StructMamPose, and its AUC is comparable to the 88.3\% of SAMA and the 88.2\% of StructMamPose.
\begin{table}[!t]
\centering
\caption{Comparison of results on MPI-INF-3DHP (ground-truth 2D input). $T$: number of input frames. PCK/AUC are in \% and MPJPE is in mm. The best and second-best displayed results are in bold and underlined, respectively. Dashes denote unreported results.}
\label{tab:mpi-comparison}
\papertablestyle
\begin{tabularx}{\columnwidth}{@{}>{\hsize=2.5\hsize\raggedright\arraybackslash}X>{\hsize=0.35\hsize\centering\arraybackslash}X>{\hsize=0.7\hsize\raggedleft\arraybackslash}X>{\hsize=0.7\hsize\raggedleft\arraybackslash}X>{\hsize=0.75\hsize\raggedleft\arraybackslash}X@{}}\toprule
\papertablehead{Method} & $T$ & \papertablehead{PCK}$\uparrow$ & \papertablehead{AUC}$\uparrow$ & \papertablehead{MPJPE}$\downarrow$ \\ \midrule
MixSTE~\cite{zhang2022mixste} {\scriptsize(CVPR'22)} & 27 & 94.4 & 66.5 & 54.9 \\
P-STMO~\cite{PSTMO2022} {\scriptsize(ECCV'22)} & 81 & 97.9 & 75.8 & 32.2 \\
GLA-GCN~\cite{GLAGCN2023} {\scriptsize(ICCV'23)} & 81 & 98.5 & 79.1 & 27.8 \\
STCFormer~\cite{STCFormer2023} {\scriptsize(CVPR'23)} & 81 & 98.7 & 83.9 & 23.1 \\
MotionAGFormer-L~\cite{Mehraban2024MotionAGFormer} {\scriptsize(WACV'24)} & 81 & 98.2 & 85.3 & 16.2 \\
KTPFormer~\cite{KTPFormer2024} {\scriptsize(CVPR'24)} & 81 & 98.9 & 85.9 & 16.7 \\
SCT~\cite{Zheng2026SCT} {\scriptsize(PR'26)} & 81 & 98.7 & 86.6 & 16.0 \\
TCPFormer~\cite{TCPFormer2025} {\scriptsize(AAAI'25)} & 81 & \underline{99.0} & 87.7 & 15.0 \\
Pose Magic~\cite{PoseMagic2025} {\scriptsize(AAAI'25)} & 81 & 98.8 & 87.6 & 14.7 \\
SDFormer-B~\cite{Zhong2026SDFormer} {\scriptsize(PR'26)} & 81 & 98.9 & 88.0 & 14.6 \\
PoseMamba-B~\cite{huang2025posemamba} {\scriptsize(AAAI'25)} & 81 & --- & --- & 14.5 \\
SAMA~\cite{lu2025sama} {\scriptsize(ICCV'25)} & 81 & \underline{99.0} & \textbf{88.3} & 14.4 \\
StructMamPose~\cite{jiang2026structmampose} {\scriptsize(ICML'26)} & 81 & 98.9 & \underline{88.2} & \textbf{14.1} \\
\textbf{SCR-Mamba-X} & 81 & \textbf{99.3} & 88.1 & \underline{14.2} \\ \bottomrule
\end{tabularx}
\end{table}

\subsection{Computational cost and inference speed}
\label{sec:efficiency}
Since different papers count costs differently, mainly in selective scanning and the functional projection parameters, we take the public implementation of PoseMamba as a reference and count the parameters and the computational cost of a single forward pass under the same rule (Table~\ref{tab:common-count}). Selective scanning is converted into MACs according to the cost given by the Mamba authors\footnote{The scan cost given by the \href{https://github.com/state-spaces/mamba/issues/110}{Mamba authors} is $9dN$ floating-point operations per step, which we convert by assuming that 1 MAC corresponds to 2 floating-point operations.}, and the 2D detector and flip testing are not counted.

Compared with PoseMamba-X, SCR-Mamba-L achieves a lower P1 with 76.8\% fewer parameters and 90.3\% lower computational cost. The reduction in computational cost mainly comes from the hierarchical structure, which places the main modeling blocks at the 81-frame and 27-frame scales.

\begin{table}[!t]
\centering

\caption{Model cost and accuracy under the unified counting rule. The parameters are counted over the whole network, and the MACs of a single forward pass include selective scanning; the values in parentheses denote the reductions relative to PoseMamba of the corresponding scale. P1 is taken from Table~\ref{tab:main-comparison}.}
\label{tab:common-count}
\papertablestyle
\begin{tabularx}{\columnwidth}{@{}X>{\hspace{0.9em}}r>{\hspace{0.9em}}r>{\hspace{0.9em}}r@{}}\toprule
\papertablehead{Model} & \papertablehead{Params (M)} & \papertablehead{MACs (G)} & \papertablehead{P1} $\downarrow$ \\ \midrule
PoseMamba-S & 1.47 & 8.26 & 41.8 \\
\textbf{SCR-Mamba-S} & 1.09 {\scriptsize($-25.9\%$)} & 2.21 {\scriptsize($-73.2\%$)} & 39.2 \\
\addlinespace[3pt]
PoseMamba-B & 4.74 & 23.96 & 40.8 \\
\textbf{SCR-Mamba-B} & 3.87 {\scriptsize($-18.4\%$)} & 11.11 {\scriptsize($-53.6\%$)} & 37.8 \\
\addlinespace[3pt]
PoseMamba-L & 9.45 & 47.93 & 38.1 \\
\textbf{SCR-Mamba-L} & 7.72 {\scriptsize($-18.3\%$)} & 15.10 {\scriptsize($-68.5\%$)} & 36.8 \\
\addlinespace[3pt]
PoseMamba-X & 33.31 & 155.42 & 37.1 \\
\textbf{SCR-Mamba-L} & 7.72 {\scriptsize($-76.8\%$)} & 15.10 {\scriptsize($-90.3\%$)} & 36.8 \\
\textbf{SCR-Mamba-X} & 16.62 {\scriptsize($-50.1\%$)} & 32.05 {\scriptsize($-79.4\%$)} & 36.6 \\
\bottomrule
\end{tabularx}
\end{table}

\paragraph{Inference speed}
Table~\ref{tab:inference-speed} reports the inference time on a single RTX 4090 GPU. With a batch size of 1, SCR-Mamba-L takes 12.2\,ms, about 2.3$\times$ and 1.4$\times$ faster than PoseMamba-L and MotionBERT, respectively. With a batch size of 32, its per-sequence time decreases to 5.3\,ms, and the throughput is about 5.2$\times$ and 3.0$\times$ higher than those of the two methods.

\begin{table}[!t]
\centering
\caption{Per-sequence inference time (ms) and P1 (mm) on a single RTX 4090 GPU. The input contains 243 frames, and 2D detection and flip testing are excluded; for a batch size of 32, the time is the whole-batch time divided by 32.}
\label{tab:inference-speed}
\papertablestyle
\begin{tabularx}{\columnwidth}{@{}X>{\hspace{0.9em}}r>{\hspace{0.9em}}r>{\hspace{0.9em}}r@{}}\toprule
\papertablehead{Model} & \papertablehead{Batch 1} & \papertablehead{Batch 32} & \papertablehead{P1}$\downarrow$ \\ \midrule
MotionBERT~\cite{zhu2023motionbert} & 17.3 & 15.7 & 39.2 \\
PoseMamba-L~\cite{huang2025posemamba} & 28.6 & 27.5 & 38.1 \\
\textbf{SCR-Mamba-S} & 5.8 & 1.3 & 39.2 \\
\textbf{SCR-Mamba-B} & 6.8 & 3.6 & 37.8 \\
\textbf{SCR-Mamba-L} & 12.2 & 5.3 & 36.8 \\
\textbf{SCR-Mamba-X} & 13.6 & 8.4 & 36.6 \\
\bottomrule
\end{tabularx}
\end{table}

\subsection{Ablation studies}
\label{sec:ablation-studies}
We conduct ablation studies on Human3.6M based on SCR-Mamba-L, and successively analyze the overall effects of SCSM and CSIR, the temporal hierarchy, the receiving branches and conditioned targets of the skeleton context, the cross-scale fusion form, and consistent transfer. Each modeling block contains one step of parallel spatial and temporal modeling and a feed-forward network. All configurations use the same channel width, total number of modeling blocks, and training settings.

\begin{figure*}[!t]
\centering
\includegraphics[width=0.92\linewidth]{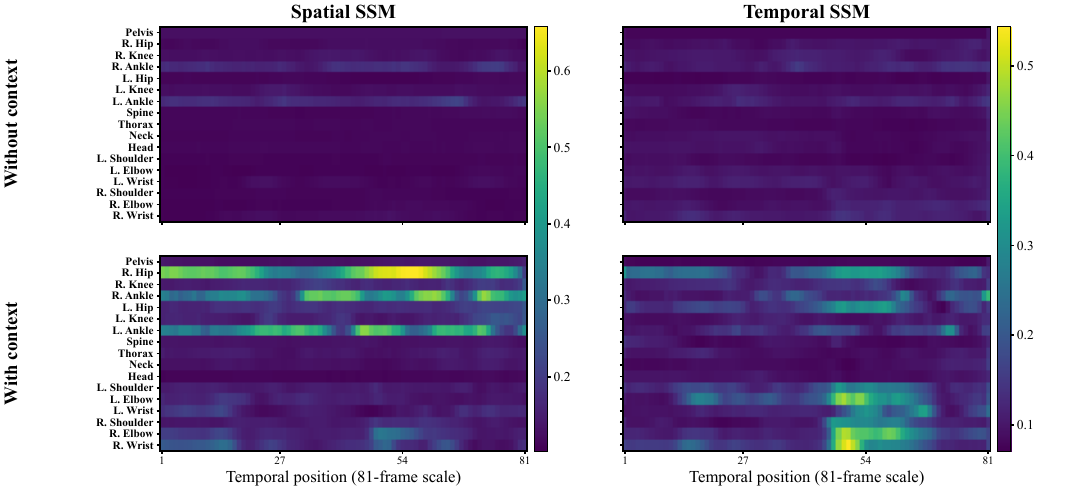}
\caption{Effect of the skeleton context on the state update step size. With the model and input features fixed, the average step size $\Delta$ of each joint in the 9th modeling block (81-frame scale) on a Walking clip is shown without and with the skeleton context $K$, and the two maps of each branch share the same color scale.}
\label{fig:scsm-heatmap}
\end{figure*}

\begin{figure*}[!t]
\centering
\includegraphics[width=0.95\linewidth]{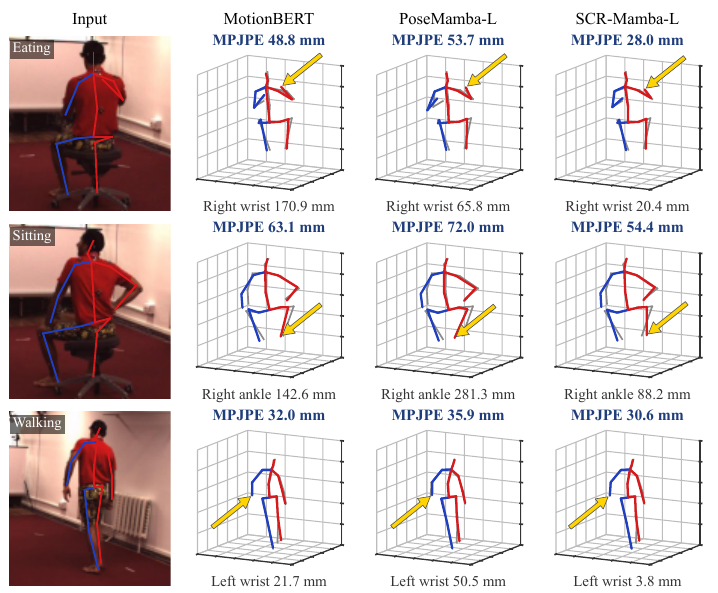}
\caption{Qualitative comparisons with MotionBERT and PoseMamba-L on Human3.6M. Blue and red denote the predicted left and right limbs, respectively, and gray denotes the ground truth; the input column shows the SH 2D detections. Each panel is annotated with the whole-body MPJPE, and the 3D error (mm) of the joint indicated by the yellow arrow is given below it.}
\label{fig:qualitative-comparison}
\end{figure*}

\begin{figure*}[!t]
\centering
\includegraphics[width=0.98\linewidth]{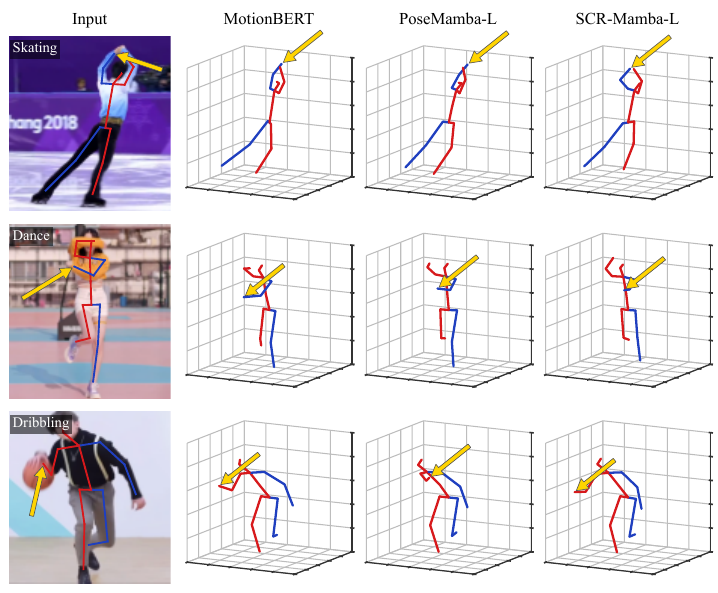}
\caption{Qualitative comparisons with MotionBERT and PoseMamba-L on public demo videos. Blue and red denote the left and right limbs, respectively, the input column shows the 2D joints detected by RTMPose, and the yellow arrows indicate the joints with large differences among the methods.}
\label{fig:wild-comparison}
\end{figure*}

\paragraph{Effect of SCSM and CSIR}
We first evaluate the effects of SCSM and CSIR, and the results are shown in Table~\ref{tab:core-ablation}. A0 performs parallel spatial and temporal state modeling on the full-length sequence, with the selective parameters generated from the features of each branch; A1 adds SCSM, A2 introduces CSIR, and A3 adopts both, i.e., SCR-Mamba-L.

\begin{table}[!t]
\centering
\caption{Ablation results of SCSM, CSIR, and the temporal hierarchy. The CSIR column indicates the number of temporal levels, and each configuration has a channel width of 128 and 20 modeling blocks in total. P1/P2 are in mm.}
\label{tab:core-ablation}
\papertablestyle
\setlength{\tabcolsep}{4pt}\begin{tabularx}{\columnwidth}{@{}>{\hsize=1.2\hsize\raggedright\arraybackslash}X>{\hsize=0.95\hsize\centering\arraybackslash}X>{\hsize=1.0\hsize\centering\arraybackslash}X>{\hsize=1.1\hsize\raggedleft\arraybackslash}X>{\hsize=1.0\hsize\raggedleft\arraybackslash}X>{\hsize=0.875\hsize\raggedleft\arraybackslash}X>{\hsize=0.875\hsize\raggedleft\arraybackslash}X@{}}\toprule
\papertablehead{Config.} & \papertablehead{SCSM} & \papertablehead{CSIR} & \papertablehead{Params (M)} & \papertablehead{MACs (G)} & \papertablehead{P1}$\downarrow$ & \papertablehead{P2}$\downarrow$ \\ \midrule
A0 & No & No & 7.22 & 37.86 & 39.7 & 31.6 \\
A1 & Yes & No & 7.72 & 39.98 & 37.4 & 31.1 \\
A2 & No & 2-level & 7.22 & 14.30 & 37.8 & 31.8 \\
A3$'$ & Yes & 1-level & 7.72 & 18.66 & 36.8 & 30.9 \\
\textbf{A3} & Yes & 2-level & 7.72 & 15.10 & 36.8 & 31.1 \\ \bottomrule
\end{tabularx}
\end{table}

Compared with A0, adding SCSM (A1) reduces P1/P2 by 2.3/0.5\,mm, and introducing CSIR (A2) reduces P1 by 1.9\,mm while lowering the computational cost from 37.86\,G to 14.30\,G. When both are adopted, the P1 of A3 is 0.6\,mm lower than that of A1. Meanwhile, the computational cost is reduced from 39.98\,G to 15.10\,G, i.e., by 62.2\%, with only 512 additional aggregation and interpolation parameters. In the hierarchical structure, SCSM reduces P1 from 37.8\,mm to 36.8\,mm.

\paragraph{Effect of the temporal hierarchy}
Table~\ref{tab:core-ablation} also compares the full-length (A1), single-level (A3$'$), and two-level (A3) temporal structures, all of which adopt SCSM. The single-level and two-level structures both achieve a P1 of 36.8\,mm, lower than that of the full-length configuration. At a similar P1, the two-level structure reduces the computational cost from 18.66\,G to 15.10\,G, i.e., by 19.1\%, while P2 increases from 30.9\,mm to 31.1\,mm.

\paragraph{Effect of the skeleton context}
Table~\ref{tab:structure-ablation} analyzes the receiving branches and relation types of the skeleton context. When only the spatial branch receives the context, P1 is reduced by 0.6\,mm compared with the configuration without context, and by 0.1\,mm when only the temporal branch receives it, indicating that the structural guidance in the spatial branch contributes more. When both branches receive the context, P1/P2/Distal are further reduced by 0.4/0.2/0.6\,mm compared with the spatial-only configuration, and the P1 reduction of 1.0\,mm exceeds the sum of the two single-branch reductions, indicating that the structural guidance of the two branches is complementary. Removing the symmetry relations increases its P1/P2 to 37.6/31.6\,mm, which demonstrates that the left--right limb relations provide effective information beyond bone connections.

SCSM also reduces the errors of distal joints. The average error of SCR-Mamba-L on both wrists and ankles is 55.4\,mm, which is 1.8\,mm lower than that of the configuration without context (A2). Compared with A2, SCR-Mamba-L reduces the errors of all non-root joints, with reductions of 2.8, 2.0, and 1.6\,mm on the left ankle, right wrist, and right elbow, respectively.

To observe the effect of SCSM on state updates, we fix the trained model and input features, and compare the state update step size $\Delta$ of each modeling block with and without the skeleton context $K$. Over all 20 modeling blocks on three test clips, $K$ changes the step sizes of the spatial and temporal branches by 0.077 and 0.040 on average, which is consistent with the larger gain of the spatial branch in Table~\ref{tab:structure-ablation}. In most modeling blocks, retaining $K$ increases both the frame-wise fluctuation and the inter-joint difference of the step sizes, indicating that the skeleton context makes state updates vary more evidently across joints and time steps. Fig.~\ref{fig:scsm-heatmap} takes the 9th modeling block as an example: with $K$ retained, the step sizes of the right hip and both ankles in the spatial branch increase markedly, and those of the elbows and wrists in the temporal branch increase during some periods.

\begin{table}[!t]
\centering
\caption{Receiving branches and relation types of the skeleton context. Spatial/Temporal: the branches receiving the context. Bone/sym.: bone/symmetry relations. Distal: the average position error of both wrists and ankles (mm).}
\label{tab:structure-ablation}
\papertablestyle
\begin{tabularx}{\columnwidth}{@{}>{\hsize=1.7\hsize\raggedright\arraybackslash}X>{\hsize=0.9\hsize\centering\arraybackslash}X>{\hsize=1.1\hsize\centering\arraybackslash}X>{\hsize=0.7\hsize\raggedleft\arraybackslash}X>{\hsize=0.7\hsize\raggedleft\arraybackslash}X>{\hsize=0.9\hsize\raggedleft\arraybackslash}X@{}}\toprule
\papertablehead{Relation} & \papertablehead{Spatial} & \papertablehead{Temporal} & \papertablehead{P1}$\downarrow$ & \papertablehead{P2}$\downarrow$ & \papertablehead{Distal}$\downarrow$ \\ \midrule
None & No & No & 37.8 & 31.8 & 57.2 \\
Bone + sym. & Yes & No & 37.2 & 31.3 & 56.0 \\
Bone + sym. & No & Yes & 37.7 & 31.7 & 56.5 \\
Bone & Yes & Yes & 37.6 & 31.6 & 57.1 \\
\textbf{Bone + sym.} & Yes & Yes & 36.8 & 31.1 & 55.4 \\ \bottomrule
\end{tabularx}
\end{table}

\paragraph{Conditioned targets of the skeleton context}
Table~\ref{tab:scsm-target} compares different conditioned targets of the skeleton context $K$. When $K$ is directly added to the features, which corresponds to feature-level fusion, it enters the scan content, output gate, and selective parameters simultaneously. Jointly conditioning $\Delta_a$, $\mathbf B_a$, and $\mathbf C_a$ achieves the lowest P1, which is 0.5\,mm lower than directly adding $K$ to the features, and 0.6 and 0.1\,mm lower than conditioning only $\Delta_a$ or only $\mathbf B_a$ and $\mathbf C_a$, respectively. This indicates that jointly regulating the step size and state writing and reading with the same skeleton context is more effective than fusing it into the features or regulating only a subset of the parameters.

\begin{table}[!t]
\centering
\caption{Ablation on the conditioned targets of the skeleton context. \checkmark{} indicates that the component is generated from $N_a+K$, and otherwise only from $N_a$. Content includes the scan content and output gate. P1/P2 are in mm.}
\label{tab:scsm-target}
\papertablestyle
\begin{tabularx}{\columnwidth}{@{}>{\hsize=1.6\hsize\raggedright\arraybackslash}X>{\hsize=0.6\hsize\centering\arraybackslash}X>{\hsize=0.8\hsize\centering\arraybackslash}X>{\hsize=1.0\hsize\centering\arraybackslash}X>{\hsize=0.7\hsize\raggedleft\arraybackslash}X>{\hsize=0.7\hsize\raggedleft\arraybackslash}X@{}}\toprule
\papertablehead{Target} & $\Delta_a$ & $\mathbf B_a,\mathbf C_a$ & \papertablehead{Content} & \papertablehead{P1}$\downarrow$ & \papertablehead{P2}$\downarrow$ \\ \midrule
None & -- & -- & -- & 37.8 & 31.8 \\
Feature & \checkmark & \checkmark & \checkmark & 37.3 & 31.0 \\
$\Delta_a$ only & \checkmark & -- & -- & 37.4 & 31.4 \\
$\mathbf B_a,\mathbf C_a$ only & -- & \checkmark & -- & 36.9 & 31.1 \\
\textbf{\boldmath$\Delta_a,\mathbf B_a,\mathbf C_a$} & \checkmark & \checkmark & -- & 36.8 & 31.1 \\ \bottomrule
\end{tabularx}
\end{table}

\paragraph{Effect of the cross-scale fusion form}
Table~\ref{tab:return-form} compares three fusion forms: $P_cG$ directly recovers the fine-scale representation from the coarse-scale features, corresponding to the recovery-based design in Fig.~\ref{fig:intro-concept}(b)(i); $Z+P_cG$, which is the widely used skip-connection fusion, adds the retained features to the complete coarse-scale features; and $Z+P_c(G-D_cZ)$ adds only the increments produced by coarse-scale modeling. The incremental update achieves a P1 of 36.8\,mm, which is 0.9 and 1.2\,mm lower than those of the former two forms, respectively, and its MPJVE also decreases from 1.70 and 1.68 to 1.66\,mm/frame. Since $G=D_cZ+\delta$, $Z+P_cG$ additionally superimposes the smoothed fine-scale features $P_cD_cZ$, which explains why it performs even worse than direct recovery. Compared with adding the complete features, the incremental update reduces the errors of the thorax, head, and right ankle by 4.9, 2.4, and 2.1\,mm, respectively.

\begin{table}[!t]
\centering
\caption{Comparison of cross-scale fusion forms. $Z$: fine-scale feature. $G$: coarse-scale modeling output. The MACs of all three configurations are 15.10\,G. P1/P2 are in mm, and MPJVE is in mm/frame.}
\label{tab:return-form}
\papertablestyle
\begin{tabularx}{\columnwidth}{@{}>{\hsize=2.1\hsize\raggedright\arraybackslash}X>{\hsize=0.6\hsize\raggedleft\arraybackslash}X>{\hsize=0.6\hsize\raggedleft\arraybackslash}X>{\hsize=0.7\hsize\raggedleft\arraybackslash}X@{}}\toprule
\papertablehead{Update} & \papertablehead{P1}$\downarrow$ & \papertablehead{P2}$\downarrow$ & \papertablehead{MPJVE}$\downarrow$ \\ \midrule
$P_cG$ & 37.7 & 31.7 & 1.70 \\
$Z+P_cG$ & 38.0 & 31.2 & 1.68 \\
\textbf{\boldmath$Z+P_c(G-D_cZ)$} & 36.8 & 31.1 & 1.66 \\ \bottomrule
\end{tabularx}
\end{table}

\paragraph{Effect of consistent transfer}
Table~\ref{tab:transfer-ablation} examines the combined effect of content-adaptive transfer and consistent transfer. Using learned weights without consistent transfer leads to the highest errors. After adding consistent transfer, P1/P2 are reduced by 0.3/0.1\,mm, P1 becomes 0.2\,mm lower than that with fixed weights, and the two-level re-aggregation residuals decrease from the order of $10^{-3}$ to the order of $10^{-8}$. Therefore, when the content-adaptive weights are used together with consistent transfer, the model achieves the lowest P1 and ensures consistent updates across the two scales.

\begin{table}[!t]
\centering
\caption{Ablation on content-adaptive transfer and consistent transfer. Fixed/Learned: fixed or learnable aggregation and interpolation weights. Consistency: whether consistent transfer is used. P1/P2 are in mm.}
\label{tab:transfer-ablation}
\papertablestyle
\begin{tabularx}{\columnwidth}{@{}>{\hsize=1.5\hsize\raggedright\arraybackslash}X>{\hsize=1.2\hsize\centering\arraybackslash}X>{\hsize=0.65\hsize\raggedleft\arraybackslash}X>{\hsize=0.65\hsize\raggedleft\arraybackslash}X@{}}\toprule
\papertablehead{Transfer} & \papertablehead{Consistency} & \papertablehead{P1}$\downarrow$ & \papertablehead{P2}$\downarrow$ \\ \midrule
Fixed & No & 37.0 & 31.1 \\
Learned & No & 37.1 & 31.2 \\
\textbf{Learned} & Yes & 36.8 & 31.1 \\ \bottomrule
\end{tabularx}
\end{table}

\subsection{Qualitative analysis}
\label{sec:additional-experiments}

\paragraph{Human3.6M}
Fig.~\ref{fig:qualitative-comparison} presents a qualitative comparison of MotionBERT, PoseMamba-L, and SCR-Mamba-L. We select frames with clear improvements on distal joints from the Eating, Sitting, and Walking sequences of S9. SCR-Mamba-L is closer to the ground truth on the distal joints indicated by the arrows. For example, in the Eating case, the right-wrist errors of MotionBERT and PoseMamba-L are 170.9 and 65.8\,mm, respectively, whereas that of SCR-Mamba-L is reduced to 20.4\,mm.

\paragraph{In-the-wild videos}
To verify the generalization ability of our approach, we directly apply the model trained on Human3.6M to public demo videos. The videos are taken from the open-source demos of PoseFormerV2, MHFormer, and GraphMLP~\cite{PoseFormerV22023,Li2022MHFormer,Li2025GraphMLP}, and include skating, dancing, and dribbling actions. The 2D joints are detected by RTMPose~\cite{Jiang2023RTMPose}. As shown in Fig.~\ref{fig:wild-comparison}, in the skating case, SCR-Mamba-L preserves the ring-shaped structure of the raised arms, whereas the arms of MotionBERT and PoseMamba-L tend to overlap in the depth direction. In the dancing case, its left arm is folded in front of the chest as in the image, whereas the left arm of MotionBERT is evidently stretched outward. In the dribbling case, its right arm reaches forward toward the basketball, whereas the right forearm of PoseMamba-L retracts toward the head.

\section{Conclusion}
\label{sec:conclusion}
This paper presents SCR-Mamba, which achieves accurate and efficient video-based 3D human pose estimation through skeleton-conditioned state modeling (SCSM) and cross-scale incremental refinement (CSIR). SCSM regulates the generation of selective parameters with the skeleton context, so that structural relations are directly involved in state updates; CSIR retains per-frame features and progressively incorporates the increments produced by coarse-scale modeling into the fine-scale representations, thereby placing the main computation on shorter sequences.

On Human3.6M, SCR-Mamba-L achieves P1/P2 errors of 36.8/31.1\,mm with 15.10\,G MACs, and SCR-Mamba-X further achieves 36.6/30.9\,mm, reaching state-of-the-art performance. On MPI-INF-3DHP, SCR-Mamba-X attains a PCK of 99.3\% and an MPJPE of 14.2\,mm. Ablation studies show that jointly regulating the selective parameters with the same skeleton context reduces P1 by 0.5\,mm compared with directly fusing the context into the features. Applying the skeleton context to both branches reduces P1 by 1.0\,mm, exceeding the sum of the single-branch reductions. Compared with fusing the complete coarse-scale features, the incremental update reduces P1 by 1.2\,mm at the same computational cost. When SCSM and CSIR are adopted together, the computational cost is reduced by 62.2\% compared with the full-length structure, while P1 is further reduced by 0.6\,mm. The incremental update form of CSIR does not depend on specific modeling operators and can serve as a reference for other hierarchical temporal models.

Nevertheless, the performance of SCR-Mamba still depends to some extent on the quality of 2D detections. Our future work will incorporate human kinematic priors to improve the robustness of the model to noisy and occluded inputs. In addition, we will further enhance the temporal guidance of the skeleton context and extend CSIR to online and multi-person 3D pose estimation.

\section*{Data availability}
The code and pretrained models will be made available upon acceptance.

\section*{Declaration of generative AI and AI-assisted technologies in the manuscript preparation process}
During the preparation of this work the authors used large language models to improve the readability and language quality of the manuscript. After using this tool/service, the authors reviewed and edited the content as needed and take full responsibility for the content of the published article.

\end{document}